\documentclass[10pt,twocolumn,letterpaper]{article}

\usepackage{cvpr}              
\usepackage[accsupp]{axessibility} 
\usepackage{graphicx}
\usepackage{caption}
\usepackage{color, colortbl}
\definecolor{grey}{rgb}{0.9,0.9,0.9}

\usepackage[dvipsnames]{xcolor}

\definecolor{cvprblue}{rgb}{0.21,0.49,0.74}
\usepackage[pagebackref,breaklinks,colorlinks,citecolor=cvprblue]{hyperref}

\def\paperID{7334} 
\def\confName{CVPR}
\def\confYear{2024}

\title{Design2Cloth: 3D Cloth Generation from 2D Masks}

\author{Jiali Zheng,
\hspace{0.4cm}
Rolandos Alexandros Potamias,
\hspace{0.4cm}
Stefanos Zafeiriou\\
Imperial College London\\
{\tt\small \{jiali.zheng, r.potamias, s.zafeirou\}@imperial.ac.uk}
}

\begin{document}

\twocolumn[{%
\renewcommand\twocolumn[1][]{#1}%
\maketitle

\begin{center}
    \centering
    \captionsetup{type=figure}
    \includegraphics[width=\textwidth,height=5.9cm]
    {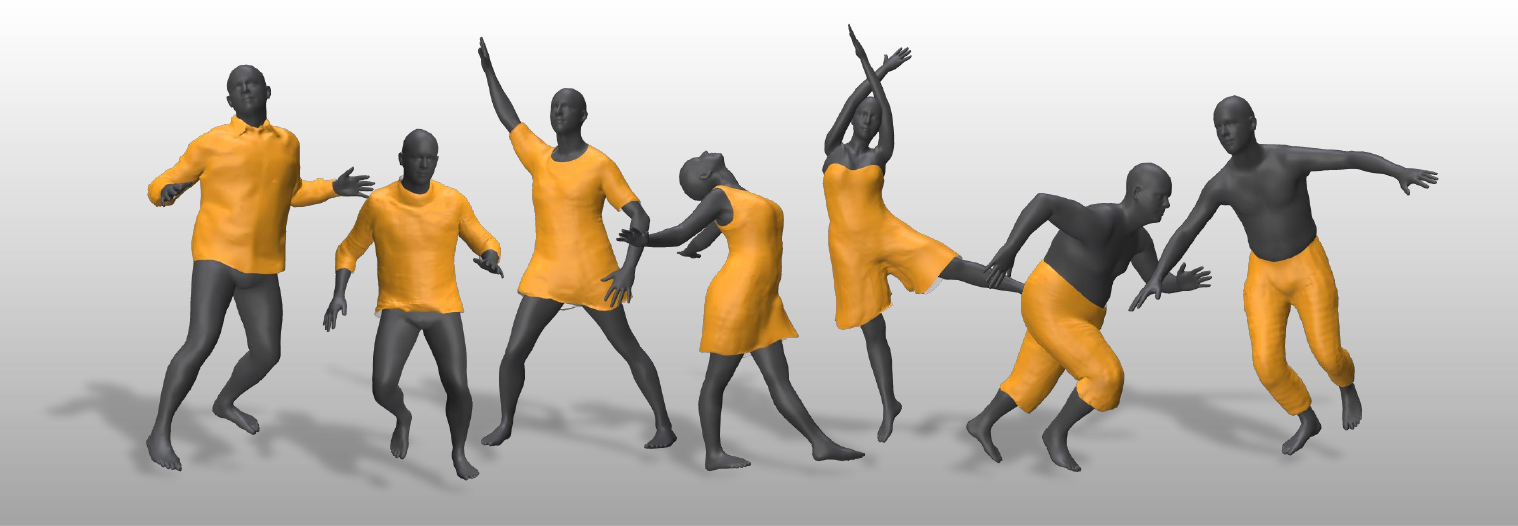}
    \captionof{figure}{We proposed Design2Cloth: a high fidelity 3D generative model for garment generation from simple 2D masks, trained on a large scale 3D cloth dataset of real-world scans. Figure illustrates interpolation between shirt (Left), trousers (Right) and shapes.}
\end{center}%
}]


\begin{abstract}
In recent years, there has been a significant shift in the field of digital avatar research, towards modeling, animating and reconstructing clothed human representations, as a key step towards creating realistic avatars. However, current 3D cloth generation methods are garment specific or trained completely on synthetic data, hence lacking fine details and realism. In this work, we make a step towards automatic realistic garment design and propose Design2Cloth, a high fidelity 3D generative model trained on a real world dataset from more than 2000 subject scans. To provide vital contribution to the fashion industry, we developed a user-friendly adversarial model capable of generating diverse and detailed clothes simply by drawing a 2D cloth mask. Under a series of both qualitative and quantitative experiments, we showcase that Design2Cloth outperforms current state-of-the-art cloth generative models by a large margin. In addition to the generative properties of our network, we showcase that the proposed method can be used to achieve high quality reconstructions from single in-the-wild images and 3D scans. Dataset, code and pre-trained model will become publicly available. \footnote{Project Page: \href{https://jiali-zheng.github.io/Design2Cloth/}{https://jiali-zheng.github.io/Design2Cloth/} }
\end{abstract}    
\section{Introduction}
\label{sec:intro}

The advent of 3D digital human avatars has facilitated the generation of realistic humans that continuously progress the gaming and filming industries \cite{SMPL:2015,handy,zheng2023ilsh}. Within this context, digitized garments emerge as a pivotal component to enhance realism. However, the diverse and perplexing nature of human garments makes modeling and generation of realistic clothed humans remains non-trivial task.  

While 2D cloth generation has been extensively studied obtaining remarkable results \cite{men2020controllable,cui2018fashiongan,cui2021dressing,honda2019viton,han2018viton,raffiee2021garmentgan,wu2021clothgan,lee2017style2vec}, the need for precise virtual try-ons necessitates the adoption of 3D cloth models that accurately simulate fabric properties, and enable more realistic digital experience. Recently, several methods have attempted to extend cloth modeling to 3D  human avatars \cite{saito2019pifu,gabeur2019moulding,zheng2021pamir}. Nevertheless, these approaches prove inadequate when it comes to accurately reconstructing garments in diverse, real-world settings. The primary cause lies in the lack of large 3D garment datasets, compared to the abundance of 2D garment datasets. In particular, real-word cloth scans \cite{Zhang_2017_CVPR,Lahner_2018_ECCV,Ma_2020_CVPR} are composed of only a few number of subjects under predefined poses wearing a limited number of garment types. To address the lack of clothed human data, several studies employed artists and simulation engines to curate synthetic data \cite{danvevrek2017deepgarment,bertiche2020cloth3d,Gundogdu_2019_ICCV, Patel_2020_CVPR,shen2020gan,Zou_2023_CVPR,zhou2023clothesnet} that cover a wider range of garment styles. 
However, synthetic cloth data have two notable inherited limitations. Firstly, simulation engines fail to accurately replicate the deformations and the wrinkles of real-world garments that leads to clothes that are characterized by an excessive degree of smoothness. Secondly, garment styles are usually generated by deforming a limited number of predefined garment categories, which constrains the model’s generalization to real-world garments. 

To overcome the limitations of generative garment models, we have collected a large-scale dataset with high resolution scans from 2010 individuals, spanning a wide range of genders, ages, heights, and weights, wearing more than 2000 unique garments. Leveraging this dataset, we trained a 3D garment generative network, that enables modeling and generation of diverse and highly detailed clothes that follow the real-world distribution. Additionally, to ease the generation process and enable a straightforward utilization of our method for individuals with no prior knowledge in the field, 
we introduce an inclusion mask representation to describe each garment. Compared to prior works that required UV maps \cite{shen2020gan,corona2021smplicit} or point clouds \cite{korosteleva2022neuraltailor,De_Luigi_2023_CVPR} to condition the garment generation, the proposed method allows users to effortlessly draw a garment mask to control the generation process.
Finally, given that the proposed model is fully differentiable, it can be used as a plug-and-play solution for inverse problems, including 3D garment reconstruction from in-the-wild images and scans.

\noindent To summarize, the main contributions of our work are:

\begin{itemize}
    \item The first large-scale cloth dataset of real-world scans, that contains more than 2k cloths from 2k subjects with various styles and body shapes.
    \item A 3D cloth generative model, 
    that is able to generate diverse garments with different types, styles and shapes. 
    Design2Cloth is founded on a user-friendly mask encoder that facilitates cloth design from 2D visibility masks. To the best of our knowledge, this is the first cloth generative model trained with over 2000 real-world cloths. 
    \item Being fully differentiable, Design2Cloth can advance the challenging task of 3D garment reconstruction, producing highly detailed 3D clothes. Our cloth reconstruction are far more realistic compared to previous state-of-the-art reconstruction methods, that fail to capture natural cloth creases and produce overly smooth results. 
    
    \item Leveraging the proposed cloth mask representation, we provide a simple but effective approach for 3D cloth reconstruction from scans. We experimentally show that the proposed method can retrieve high quality 3D garments from incomplete and corrupted inputs. 
\end{itemize}

\section{Related Work}
\label{sec:formatting}

\textbf{Generative Garment Model} for 3D clothes have emerged in recent years due to their impact in fashion design and virtual try-on applications. Early methods approached cloth modeling using Principal Component Analysis (PCA) on fixed topology garments  \cite{guan2012drape,danvevrek2017deepgarment,wang2018learning,Patel_2020_CVPR,Lahner_2018_ECCV,bhatnagar2019multi}. Nevertheless, such approaches are garment-specific i.e. they can not generalize to clothes with different templates and require training different models for each garment category.  To tackle this, Ma \etal \cite{Ma_2020_CVPR} proposed to model garments as a set of displacements on top of SMPL \cite{SMPL:2015} body topology. Following that, several methods have extended \cite{Ma_2020_CVPR}, to enrich clothes diversity and train more expressive models \cite{bertiche2020cloth3d,shen2020gan}.  
A major limitation, under this setting, is that only a limited number of cloths, that strictly follow the the contour of the body, can be represented as body displacements. 
Recently, several methods have been proposed to utilize \emph{implicit functions} to train garment agnostic generative models \cite{corona2021smplicit,Li_2022_ACCV,De_Luigi_2023_CVPR}, leveraging the property of neural implicit functions to represent topology-agnostic 3D surfaces. SMPLicit \cite{corona2021smplicit} pioneered 3D garment generative model using unsigned distance functions (UDF) to represent clothes. Most relevant to our method, DrapeNet \cite{De_Luigi_2023_CVPR} has proposed to utilize MeshUDF \cite{guillard2022meshudf} to achieve better reconstruction results. However, the aforementioned methods rely on low frequency decoders that can only generate overly smooth synthetic data that lack high frequency details, such as wrinkles and creases. In addition, they can only project clothes to the latent space of the model using point clouds and UV maps, which undoubtedly limits their applicability in real-world scenarios. Different from previous methods, we propose a high-fidelity user-friendly generative model, trained on a large scale real-word dataset, that is able to model diverse clothes with high frequency details. 
\newline

\noindent\textbf{3D Garment Datasets} have emerged as crucial resources for modeling and animating clothed human avatars. As summarized in \cref{tab:table_datasets}, existing garment datasets can be classified as either \textit{real-world} or \textit{synthetic}. The BUFF dataset \cite{Zhang_2017_CVPR} was among the first high-quality scanned datasets introduced, albeit characterized by a limited number of subjects and garments. Subsequently, several real-world datasets have emerged, with MGN \cite{bhatnagar2019multi}, DeepFashion3D \cite{zhu2020deep}, and THUman2.0 \cite{Yu_2021_CVPR} standing out as they are all composed from over 100 distinct subjects and garments. Due to the challenges associated with capturing large-scale 3D datasets of clothed humans, research efforts have been focused on the creation of synthetic cloth datasets. Leveraging physics-based simulations \cite{Patel_2020_CVPR, bertiche2020cloth3d} and artist-curated patterns \cite{shen2020gan, Zou_2023_CVPR}, synthetic datasets have successfully produced extensive collections of garments featuring diverse shapes, poses, and styles. However, they suffer from a domain gap, exhibiting overly-smoothed surfaces and non-realistic wrinkles. In this work, we take a significant step towards realistic cloth modeling by curating a large-scale real-world dataset, named DigitalMe, comprising 2010 individual subjects wearing more than 2k garments from 31 categories. 

\paragraph{Neural Invertible Skinning.}
Various approaches have been employed to tackle problems in modeling the deformation of non-rigid and articulated 3D objects such as clothed bodies. In \cite{yang2022banmo,burov2021dynamic,saito2021scanimate,wang2021metaavatar}, the Linear Blend Skinning (LBS) weights are learned separately in deformed and canonical spaces in order to utilize cycle-consistency losses for establishing correspondence. SNARF \cite{chen2021snarf} proposed to establish the correspondences using an iterative method to invert the LBS equation.  
Recently, Kant \etal \cite{kant2023invertible} proposed a pose-conditioned invertible model,  to learn a non-linear blend skinning function that preserves the correspondences with the input mesh. However, all of the aforementioned works rely on overfitting on a sequence of scans and fail to animate independent single scans. Unlike these methods, we show that using the powerful Design2Cloth encoder, we can accurately reconstruct and animate garments from single in-the-wild scans while preserving the intrinsic correspondences. 
\section{Method}
\begin{figure*}[!ht]
    \centering
    \includegraphics[width=0.9\textwidth]{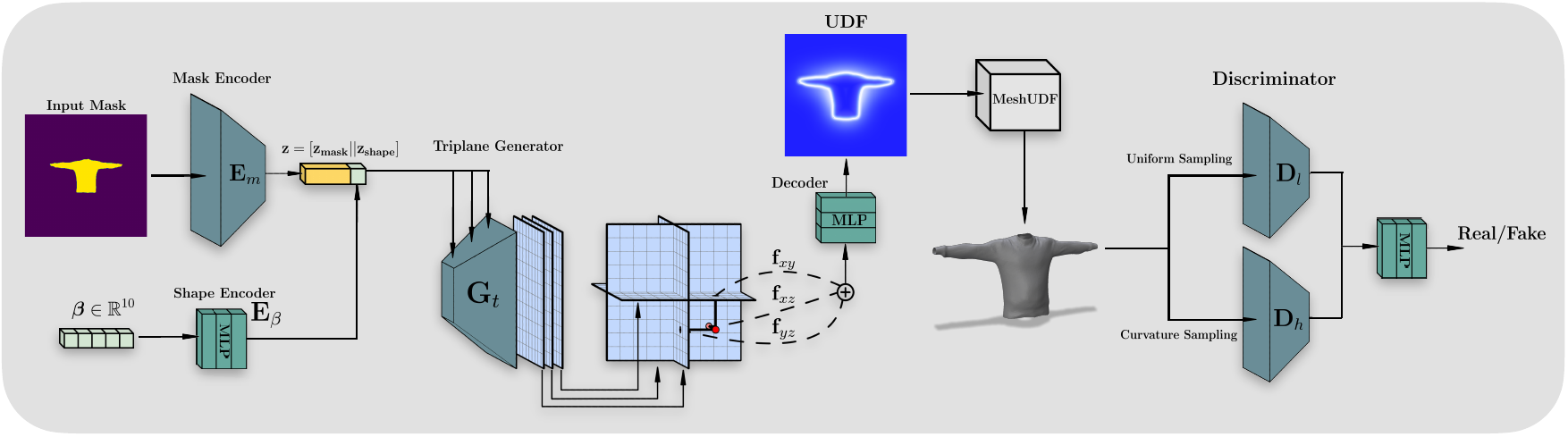}
    \vspace{-0.2cm}
    \captionof{figure}{\small{Overview of the proposed Design2Cloth: A binary mask $\mathbf{M}$ along with a shape vector $\boldsymbol{\beta}$ are fed to the encoder modules, $\textbf{E}_m, \textbf{E}_{\beta}$ to produce a latent vector $\mathbf{z}$ that is used to drive the triplane generator $\mathbf{G}_t$. The decoder network takes as input the triplane features of the projected points and regresses their corresponding unsigned distance function that is then meshed using the differentiable MeshUDF \cite{guillard2022meshudf}. To enforce the generation of highly detailed clothes we utilize a dual resolution discriminator network $\mathcal{D}$, that take as input two sparse point clouds sampled from the surface of the generated cloth. The low frequency branch $\textbf{D}_l$ takes as input a uniformly sampled point cloud whereas the high frequency branch $\textbf{D}_h$ takes as input a point cloud sampled from coarse areas of the garment surface.}
    \label{fig:method}}
\end{figure*}
\subsection{Large Scale Cloth Dataset}
\label{sec:dataset}

To train a high resolution 3D cloth generative model, we collected a large dataset, namely DigitalMe,  comprising of high resolution 3D clothed human scans. 
The scans were captured using a 3dMD multi-camera structured light stereo system, with 14 cameras and 12 uniform lighting LED panels. We captured a total of 2010 distinct subjects spanning different ages, body types and ethnicities. Participants wore a variety of garments and were scanned in numerous poses, starting from the canonical pose. The raw scans have a resolution of approximately 150,000 vertices. 

\newcolumntype{g}{>{\columncolor{Gray}}c}
\begin{table}
\begin{center}
\caption{Comparison of existing 3D garment methods. Methods that utilize \colorbox{grey}{real-world data} are highlighted in grey, while the rest are based on synthetic data. \#Garment: denotes the number of different individual garments, \#Subjects: the number of individual identities, \#Categories: the number of distinct cloth types,  Generation: the generative ability of the network, Garment-Agnostic: whether the network can represent different cloth categories.}
\label{tab:table_datasets}
\scalebox{0.58}{
\begin{tabular}{l c c c c c c c c c c}
\hline \hline

Method &\#Garment & \#Subjects & \#Categories  & Generation & 
        Garment-Agnostic  \\
\hline 

DRAPE \cite{guan2012drape}  & 420 & - & 5  & \checkmark & $\times$  \\

DeepGarment~\cite{danvevrek2017deepgarment}  & 
        2 & - & 2  & $\checkmark$ & 
        $\times$  \\

Wang \etal \cite{wang2018learning} & 8k & - & 3  & $\checkmark$ & $\times$  \\


Santesteban \etal \cite{santesteban2019learning} & 17 & - & 1   & $\checkmark$ & $\times$ \\

GarNet \cite{Gundogdu_2019_ICCV} & 1.5k & - & 3  & $\checkmark$ & $\times$  \\

TailorNet \cite{Patel_2020_CVPR}  & 207 & - & 4   & $\checkmark$ & $\times$  \\

Cloth3D \cite{bertiche2020cloth3d} & 11.3k & - & 7  & $\checkmark$ & $\checkmark$  \\

Shen \etal \cite{shen2020gan} & 104 & - & 5   & $\checkmark$ & $\checkmark$  \\

SMPLicit \cite{corona2021smplicit} & 2M & - & 11  & $\checkmark$ & $\checkmark$  \\

DIG \cite{Li_2022_ACCV}  &200 &- &2  & $\checkmark$ & $\checkmark$ \\

DrapeNet~\cite{De_Luigi_2023_CVPR}  &
        900 & - & 5& $\checkmark$ & 
        $\checkmark$ \\

HOOD~\cite{Grigorev_2023_CVPR}  & 
        15 & - & 7   & $\times$ & 
        $\checkmark$ \\

Zhao \etal \cite{Zhao_2023_CVPR}  & 
        2 & - & 2  & $\times$ & 
        $\times$ \\

CLOTH4D \cite{Zou_2023_CVPR}  & 1k &- & 6 & $\times$ & - \\

ClothesNet \cite{zhou2023clothesnet}  & 4.4k & - & 11  & $\times$ & - \\ \hline

\rowcolor{grey}
BUFF \cite{Zhang_2017_CVPR} & 24 &6 &4  & $\times$ & $\checkmark$  \\

\rowcolor{grey}
ClothCap~\cite{pons2017clothcap} & 
        - & - & -   & $\times$ & 
        $\times$ \\

\rowcolor{grey}
DeepWrinkles \cite{Lahner_2018_ECCV} & 4 &2 &2 & $\checkmark$ & $\times$  \\

\rowcolor{grey}
MGN \cite{bhatnagar2019multi}  & 712 & 356 & 5  & $\checkmark$ & $\times$ \\

\rowcolor{grey}
CAPE \cite{Ma_2020_CVPR}  & 8 &11 & 5  & $\checkmark$ &$\checkmark$  \\

\rowcolor{grey}
DeepFashion3D \cite{zhu2020deep}  & 563 & - & 10   &$\times$& $\checkmark$  \\

\rowcolor{grey}
THUman2.0 \cite{Yu_2021_CVPR}  & - &200 & -   & $\times$ & $\checkmark$ \\

\rowcolor{grey}
LGN \cite{Aggarwal_2022_ACCV}  &140 &140 &5  & $\times$ &$\checkmark$  \\

\rowcolor{grey}
GarmCap \cite{Lin_2023_ICCV} & 
        4 & - & 3   & $\times$ & 
        $\times$   \\

\rowcolor{grey}
\hline
\textbf{Design2Cloth}  & \textbf{2k}
     & \textbf{2k}  &  31    &$\checkmark$
         &$\checkmark$   \\

\hline \hline
\end{tabular}}
\end{center}
\end{table}
Using an automated pipeline we fitted parametric SMPL \cite{SMPL:2015} model and extracted the 3D clothes from the raw human scans.  Specifically, we rendered each scan from 40 different views that span $360^o$ degrees around the subject. For each view, we utilized MediaPipe algorithm \cite{bazarevsky2020blazepose} to extract 2D joint locations and subsequently lift them to 3D using linear triangulation. 
 To fit SMPL parametric body model \cite{SMPL:2015} to the 3D joint location, we optimized pose $\boldsymbol{\theta}$ and shape parameters $\boldsymbol{\beta}$ of SMPL, by minimizing the joint error $\mathcal{L}_{J}$ and the Chamfer Distance (CD) $\mathcal{L}_{cd}$ between the body template and the scan. Additionally, to enforce the SMPL body fittings to lie under the raw scan surface, an additional loss function was employed to penalize the SMPL vertices located outside of the scan.  
To extract the 3D cloth meshes from the scans we initially segmented each rendering using SAM \cite{kirillov2023segany} and then projected the 3D mesh to each one of the renderings. 
Using a majority vote, between all 40 views, we identify the vertices that correspond to cloth regions. 
 The cloth vertices were then cropped from the raw scan and canonicalized using inverse linear blend skinning function (LBS$^{-1}$). Note that the canonicalization is performed only in the pose space to retain shape identity of the subject.
 After preprocessing, 2010 unique cloth meshes were generated from the scans. For additional details about the optimization pipeline we refer the reader to the supplementary material.

To build a user-friendly generative model, we represent each cloth using a 2D visibility mask. To enhance the expressivity of the model and include more cloth details, such as V-neck patterns, we constrain the visibility masks to the frontal part of the cloth by rasterizing only the vertices positioned in front of the Z-axis plane. 






\subsection{Design2Cloth: An Implicit Garment GAN}
The architecture of the proposed model is composed of three main components: the Cloth and Shape Encoders that take as input a binary visibility mask and a target shape code, respectively, and embed them into a latent representation (\cref{sec:cloth_encoder}), the Implicit Cloth Generator $\mathbf{G}_t$  that decodes the latent code to a UDF (\cref{sec:generator}) and the dual-resolution Discriminator $\mathbf{D}$ that enforces the generator to produce highly detailed garments (\cref{sec:discriminator}).  \cref{fig:method}
depicts an overview of the proposed method.

\subsubsection{Cloth Encoder}
\label{sec:cloth_encoder}
Cloth encoder can be considered as the core component of the proposed method as it learns a compact and smooth latent space which can be then used to condition the generation process and enable interpolation between different clothes. In contrast to previous methods that require complex inputs to project a cloth in the latent space, we propose visibility masks $\mathbf{M}$ as a natural representation of different garments. More specifically, both SMPLicit \cite{corona2021smplicit}, that uses an occupancy UV image, and DrapeNet \cite{De_Luigi_2023_CVPR}, that requires a 3D cloth surface to project a cloth in the latent space, apart from being impractical, fail to provide a user-friendly experience. In contrast, visibility masks are an intuitive approach that does not require any expertise knowledge and can turn the proposed method into a scalable tool for user-friendly garment design. 

The proposed cloth encoder is divided in two modules the mask encoder $\mathbf{E}_m$ and the shape encoder $\mathbf{E}_\beta$. The mask encoder $\mathbf{E}_m$ is based on a lightweight image feature extractor \cite{sandler2018mobilenetv2} which takes as input a binary mask indicating the cloth style and outputs a compact latent representation of the cloth $\mathbf{z_{mask}}$. In addition to the clothing style, we use a shape encoder $\mathbf{E}_{\beta}$ to condition the garment generation on SMPL shape parameters $\boldsymbol{\beta}$. The resulting latent vector $\mathbf{z}$ is given from the concatenation of the two modalities, style and shape, as : 
\begin{equation}
    \mathbf{z} = [\mathbf{E}_m(\mathbf{M}) || \mathbf{E}_{\beta}(\boldsymbol{\beta})]
\end{equation}
where $||$ denotes the concatenation operator. 
Using a shape conditioned encoder, our framework learn both style and shape variations of the cloths, in contrast to previous methods that can only generate clothes of the mean shape. 

\subsubsection{Implicit Cloth Generator}
\label{sec:generator}
The main objective of the generator $\mathcal{G}$ is to learn a mapping function $\mathcal{G}(\mathbf{z}) \rightarrow C \in \mathbb{R}^{N \times 3}$ from shape and style latent codes $\mathbf{z}$  to a detailed 3D cloth $C$. Given that cloth registration to a common template is a particularly expensive and challenging task \cite{bhatnagar2019multi}, we model clothes as implicit fields. 

Inspired from \cite{chan2022efficient}, we built a generative model $\mathbf{G}_t$ upon the computationally efficient and expressive hybrid tri-plane representation. The backbone of the tri-plane generator is a 2D convolutional neural network that outputs three plane feature images $H \times W \times 32$. Using this powerful representation that explicitly learns the features of the grid we can generate detailed clothes without any additional computational requirements. Specifically, using a single multi-layer perceptron (MLP) decoder we can sufficiently predict the \emph{unsigned distance} $d(\mathbf{p})$ of a point $\mathbf{p}$ to the surface of the garment. Finally, to decode the generated UDF to a discrete mesh garment, we utilized MeshUDF \cite{guillard2022meshudf} since it is fully differentiable and enables the gradient flow in our generator network. During inference, any other decoding method could be used to obtain smoother results \cite{chen2022neural,zhang2023dualmeshudf}.

\subsubsection{Dual-Resolution Discriminator}
\label{sec:discriminator}
To build a high fidelity cloth generative model that is able to reconstruct and generate highly detailed clothes we devise a novel branched discriminator $\mathcal{D}$. Following \cite{Patel_2020_CVPR} that introduced the concept of low and high level cloth generation, we propose an elevated alternative that enforces the generation of wrinkles and high frequency details through adversarial learning. More specifically, instead of using a single branch discriminator to distinguish the generated from the real samples, we divide the workload into two distinct branches that allows the separation of 
high and low level cloth details. The low-level branch $\textbf{D}_l$ takes as input a uniformly sampled garment and learns structural features of the overall shape. On the contrary, to enforce the generator to preserve the wrinkles of the input, we feed the high-frequency branch $\textbf{D}_h$ with a point cloud sampled from the surface areas with the maximum Gaussian mean curvature \cite{pc_curv}. Both high and low level branches were implemented using PointNet++ \cite{qi2017pointnet++} encoder that takes as input the sampled point cloud and aggregates their hierarchical features into a compact latent space. The concatenation of the two latent codes are then fed to a trunk network which regresses a real-fake score. Our dual frequency discrimination not only encourages the generated garment to match the distribution of real ones, but also enforces the generation of high frequency details that previous methods failed to model. 

\textbf{Training Objective:}
To train our method we use a combination of loss functions that enforce the generator $\mathcal{G}$ to accurately predict the UDF $d(\cdot)$ of randomly sampled points in the grid ($\mathcal{L}_{\textsc{UDF}}$) and generate highly detailed clothes ($\mathcal{L}^{adv}_{\mathcal{G}}$). Additionally, using the adversarial loss, we  train the discriminator network $\mathcal{D}$ to distinguish real from generated clothes ($\mathcal{L}^{adv}_{\mathcal{D}}$). Finally, to regularize the MeshUDF triangulation, we penalize the gradients of the predicted UDF to match the ground-truth gradients at each sampled point ($\mathcal{L}_{grad}$). Formally:
\begin{equation}
\begin{aligned}
    &\mathcal{L}_{\textsc{UDF}} = \sum_\mathbf{p}|| \mathcal{G}(\mathbf{z}; \mathbf{p}) - d(\mathbf{p})||_2 \\
    &\mathcal{L}^{adv}_{\mathcal{G}}  =  \mathbb{E}_{z}[\log(\mathcal{D}(\mathcal{G}(\mathbf{z}))] \\
    &\mathcal{L}^{adv}_{\mathcal{D}}  =  \mathbb{E}_{x}[\log\mathcal{D}(x)]+\mathbb{E}_{z}[\log(1-\mathcal{D}(\mathcal{G}(\mathbf{z}))] \\
    &\mathcal{L}_{grad} = \sum_\mathbf{p}|| \hat{\mathbf{g}}_\mathbf{p} - \mathbf{g}_\mathbf{p}||_2
\end{aligned}
\end{equation}
where $\mathbf{z}$ denotes a latent code produced from the encoder, $d(\mathbf{p})$ is the ground truth unsigned distance and $\mathbf{g}_\mathbf{p}, \hat{\mathbf{g}}_\mathbf{p}$ the ground truth and the predicted gradients of the UDF at a point $\mathbf{p}$. For notation purposes we omit the sampling function used in the discriminator network. 
\section{Experiments}
In this section, we conduct a comprehensive evaluation of the capabilities of the proposed Design2Cloth network to generate, interpolate, and fit `in-the-wild' clothes.


\subsection{Cloth Generation}
\label{sec:cloth_generation}
\begin{figure*}
    \centering
    \includegraphics[width=0.85\textwidth]{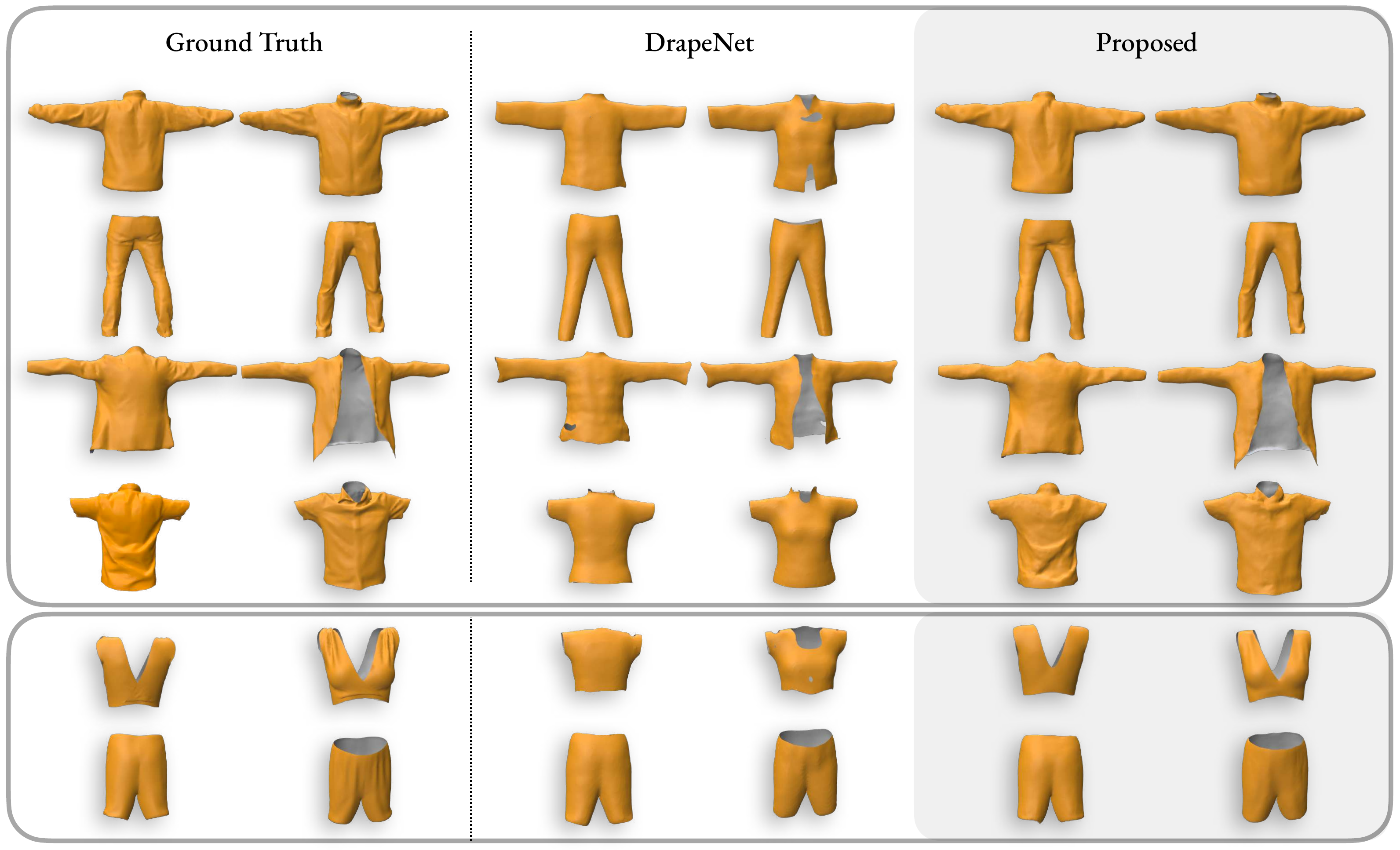}
    \captionof{figure}{\small{Qualitative comparison between the proposed and DrapeNet methods of reconstruction performance on DigitalMe (Top),  Cloth3D \cite{bertiche2020cloth3d} and ClothesNet \cite{zhou2023clothesnet} (Bottom) datasets.}
    \label{fig:reconstruction}}
\end{figure*}

\begin{table}[]
\caption{Quantitative comparison of the reconstruction performance of the Proposed and the DrapeNet \cite{De_Luigi_2023_CVPR} on the Cloth3D \cite{bertiche2020cloth3d} and DigitalMe datasets.}
\vspace{-0.5cm}
\begin{center}
\resizebox{\linewidth}{!}{
\begin{tabular}{l|cc|cc}
         & \multicolumn{2}{c|}{Cloth3D}   & \multicolumn{2}{c}{DigitalMe}        \\ \hline
Method   & CD ($\times 10^{-4}$) $\downarrow$      & NC $\uparrow$              & CD ($\times 10^{-2}$)$ \downarrow$      & NC $\uparrow$         \\ \hline
DrapeNet & 0.36          & 0.97          & 0.56          & 0.96          \\
Proposed & \textbf{0.18} & \textbf{0.99} & \textbf{0.12} & \textbf{0.98} \\ \hline \hline
\end{tabular}}
\end{center}
\label{tab:reconstruction}
\end{table}

\begin{figure}
    \centering
    \includegraphics[width=\linewidth]{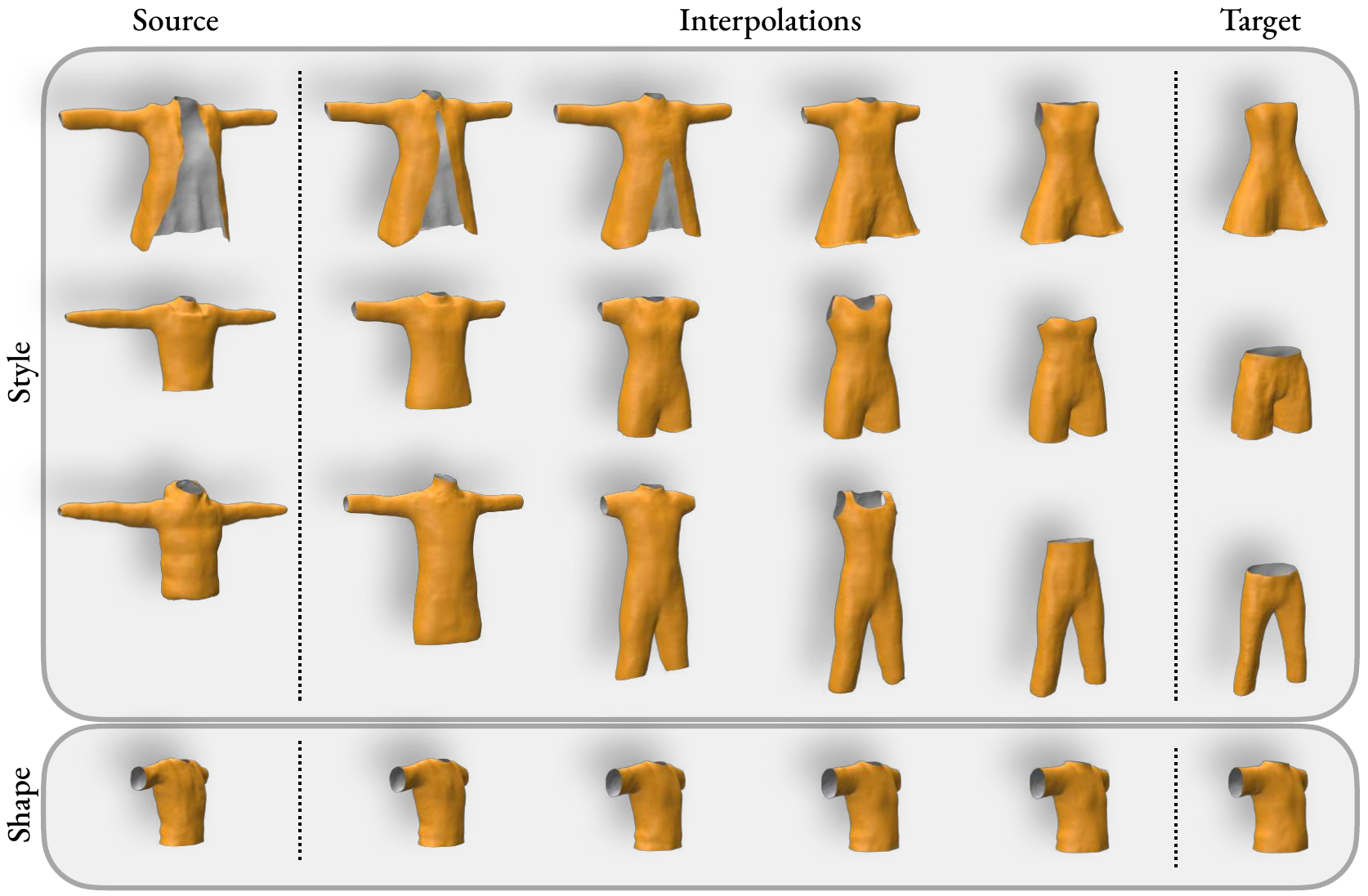}
    \vspace{-0.5cm}
    \captionof{figure}{\small{Interpolation between source and target styles (Top) and shapes (Bottom). The proposed method can interpolate between shapes and styles, generating realistic intermediate clothes. }
    \label{fig:interpolation}}
\end{figure}

To compare the performance of the proposed method on cloth generation, we trained Design2Cloth and DrapeNet \cite{De_Luigi_2023_CVPR} models on DigitalMe dataset with 100 additional clothes from  Cloth3D \cite{bertiche2020cloth3d}  ClothesNet \cite{zhou2023clothesnet}, using a common 80\%-20\% train-test split.  In \cref{tab:reconstruction}, we report the reconstruction quality of each method trained only on Cloth3D and DigitalMe datasets measured in terms of Chamfer Distance (CD) and Normal Consistency (NC) metrics. As expected, the performance of both models drop on DigitalMe dataset which contains more challenging clothes, with high frequency wrinkles and details. Despite the challenges, the proposed method not only outperforms DrapeNet under all metrics and datasets, but also manages to achieve 0.98 NC performance on DigitalMe dataset, which validates the ability of our generator to model wrinkles and creases. The superiority of the proposed method can also be qualitatively verified in \cref{fig:reconstruction}, where the proposed method can accurately generate highly detailed clothes compared to the smooth garments produced from DrapeNet. 
Furthermore, the proposed method inherits many of the desirable generative network properties such as a well-behaved latent space with smooth interpolation between diverse clothes. Specifically, as depicted in \cref{fig:interpolation}, Design2Cloth can not only smoothly interpolate between diverse garments, i.e. from an open jacket to a dress, but also transition from top to bottom clothes. Additionally, given that the proposed method is also conditioned on the cloth shape parameter, one can also interpolate between different shapes while retaining the style fixed. 

\begin{figure}
    \centering
    \includegraphics[width=\linewidth]{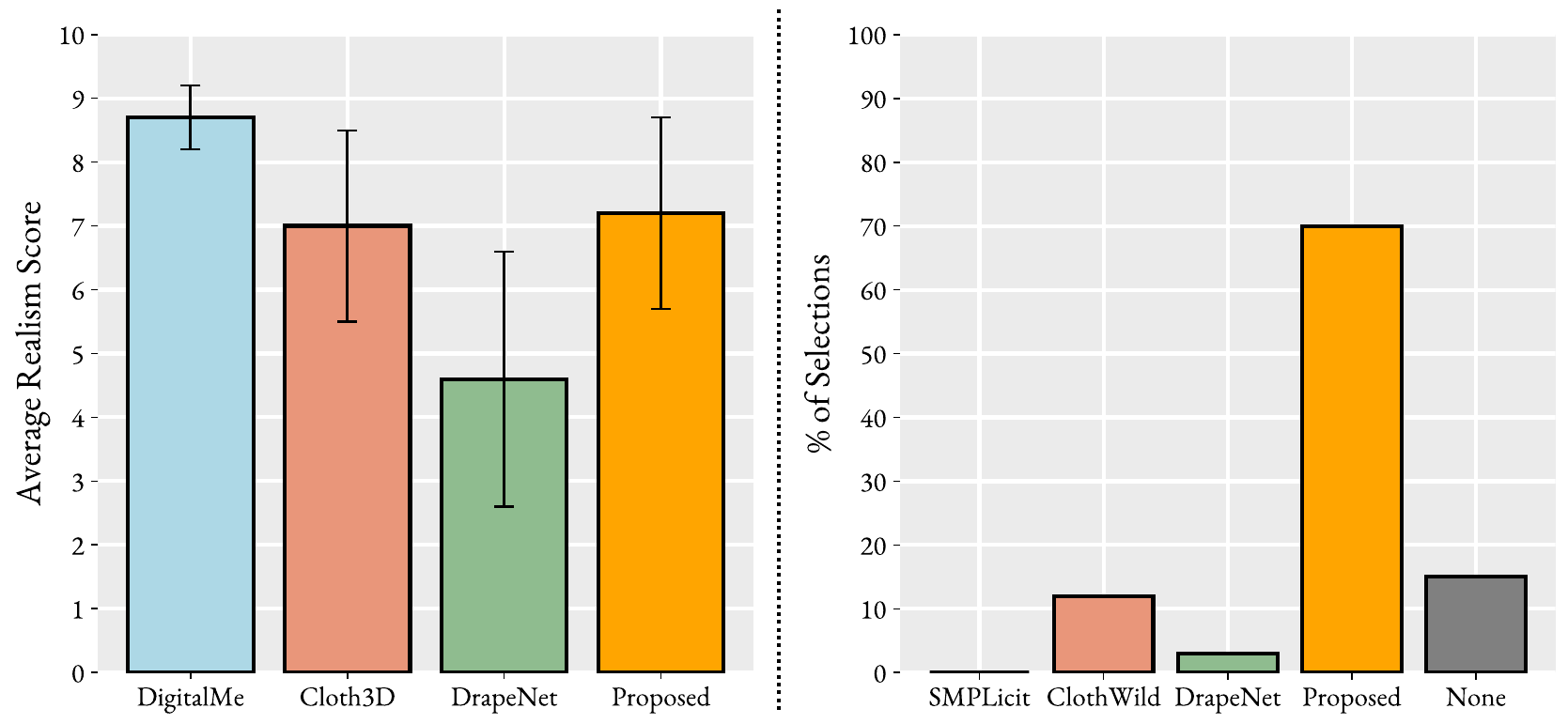}
    \captionof{figure}{\small{Human evaluation results. Left: Average realism scores of the generated and the ground truth data. Right: Perceptual Evaluation of 3D reconstructions from in-the-wild images between the proposed and the baseline methods.}
    \label{fig:user_study}}
\end{figure}

Finally, to evaluate the realism of the generated garments, we designed an online user study where 50 participants, including 20 people working in the fashion industry, were asked to assess the quality of the generations. Each participant was asked to score, on a scale from 1 (poor) to 10 (very realistic), the quality of the generated garments from the proposed and DrapeNet methods, along with the ground truth meshes from DigitalMe and Cloth3D datasets. As shown in \cref{fig:user_study} (Left), DigitalMe clothes proved to be the most realistic, achieving an average score of 8.7. Interestingly, the highly detailed clothes generated by the proposed method manages to produce more realistic clothes compared to the ground truth Cloth3D dataset and DrapeNet generations, achieving an average score of 7.2 compared to 7 and 4.6 respectively, which validates the generative power of our model. 
\subsection{3D Garment Reconstruction In-The-Wild}
Given that the proposed model is fully differentiable it can be used to fit 3D clothes from in-the-wild images. 
The first step in recovering cloth from an in-the-wild image is to estimate SMPL pose $\theta$ and shape $\beta$ parameters using an off-the-shelf pose estimation method \cite{rong2021frankmocap}. Using the detected SMPL parameters, we posed the generated cloth and projected it to the image space to obtain a posed mask. 
We optimized the latent code $\mathbf{z}$ by minimizing the Intersection over Union (IoU) between image cloth segmentation $\mathbf{S}$ extracted using SAM\cite{kirillov2023segany} and the projected posed mask.
\begin{figure}
    \centering
    \includegraphics[width=\linewidth]{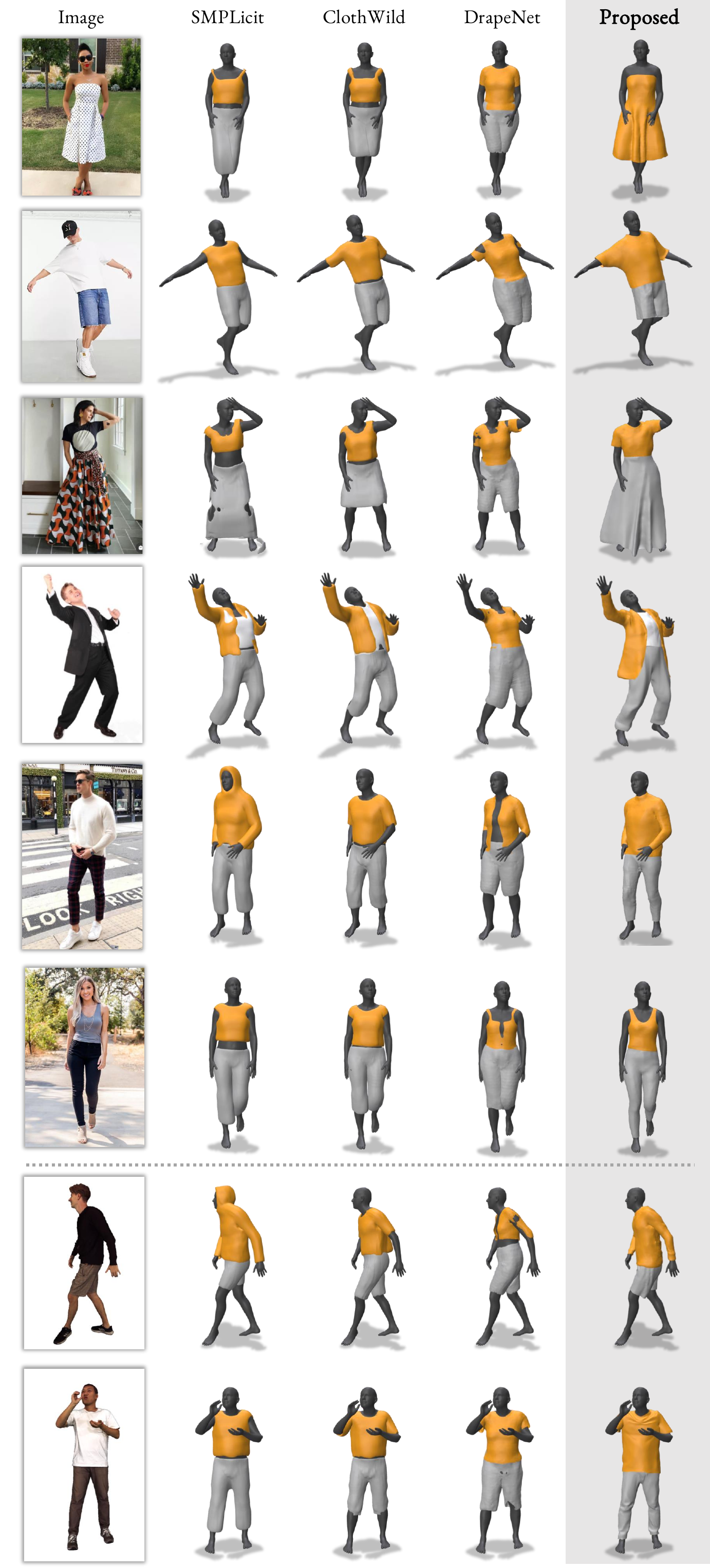}
    \vspace{-0.5cm}
    \captionof{figure}{\small{Garment reconstruction from in-the-wild images. Qualitative comparison between the proposed and the baseline method. The proposed method can generate highly detailed clothes compared to the smooth reconstructions of the baseline methods. Figure better viewed with zoom. The last two rows of the figure correspond to renderings from CustomHumans \cite{ho2023custom} dataset. }
    \label{fig:reconstruction_from_image}}
\end{figure}

To facilitate smoother convergence of the model we calculate the latent code statistics of the training set and we initialize $\mathbf{z}$ with the mean value of the dataset. Mathematically, the optimization scheme can be formulated as: 
\begin{equation}
\begin{split}
        \min_{z}\mathcal{L}_{\text{IoU}} \left(\Pi_\mathbf{M} \left(\text{LBS}(\mathcal{G}(\mathbf{z}; \boldsymbol{\beta}),\boldsymbol{\theta}) \right) , \mathbf{S}\right) + \mathcal{L}_{prior}(\mathbf{z})
\end{split}
\end{equation}
where $\Pi_\mathbf{M}$ is the 2D occupancy mask rasterized from a differentiable renderer, $\mathbf{S}$ the extracted cloth mask, $\mathcal{G}(\mathbf{z}; \boldsymbol{\beta})$ our cloth generator that maps a latent code $\mathbf{z}$ with shape $\boldsymbol{\beta}$ to a 3D cloth $C \in \mathbb{R}^{N\times 3}$, LBS$(\cdot,\boldsymbol{\theta})$ a skinning function that poses a mesh to the target pose $\boldsymbol{\theta}$ and $\mathcal{L}_{prior}(\mathbf{z})$ an $\mathcal{L}_2$ regularization that constrains $\mathbf{z}$ to feasible values.
To compare the reconstruction performance of the proposed method, we utilized three state-of-the-art 3D cloth reconstruction models, namely  SMPLicit \cite{corona2021smplicit}, ClothWild \cite{Moon_2022_ECCV_ClothWild} and DrapeNet \cite{De_Luigi_2023_CVPR}. In \cref{fig:reconstruction_from_image} we qualitatively compare the proposed and the baseline method in 3D garment reconstruction from in-the-wild images. Importantly, in contrast to the baseline methods, the proposed method requires optimization of a single network for both top and bottom garments, which significantly reduces the memory and computational requirements. As can be observed, Design2Cloth can reconstruct detailed garments, with intricate creases and puckering of fabric compared to the smooth reconstructions of the baselines. To quantitatively validate the perceptual reconstruction performance of the proposed method, we have included a second section in the human evaluation study described in \cref{sec:cloth_generation}. This time the participants were given the input image along with the reconstructions from the proposed and the baseline methods, and were asked to choose the one that has the most similarities with the input image. To avoid biases in the evaluation process, users were also given a `None' option. As can be seen in \cref{fig:user_study}, more than 70\% of the participants selected the proposed method as the one achieving the most accurate and realistic reconstructions, compared to less than 15\% that selected ClothWild method. \newline
To further quantitatively evaluate the cloth reconstruction performance of the proposed method, we utilized CustomHumans dataset \cite{ho2023custom} that contains high resolution 3D scans of dressed humans in various poses along with their corresponding SMPL fittings. Specifically, we rendered the posed scans in arbitrary views and measured the CD between the ground truth and the reconstructed garments using the optimization pipeline described above. For a fair comparison, we use the ground truth SMPL parameters. As shown in \cref{tab:customhuman_reconstruction}, the proposed method outperforms the baseline methods in CD reconstruction error by a large margin, which can be also validated from the last two rows of \cref{fig:reconstruction_from_image}.  
\begin{figure*}
    \centering
    \includegraphics[width=0.85\textwidth]{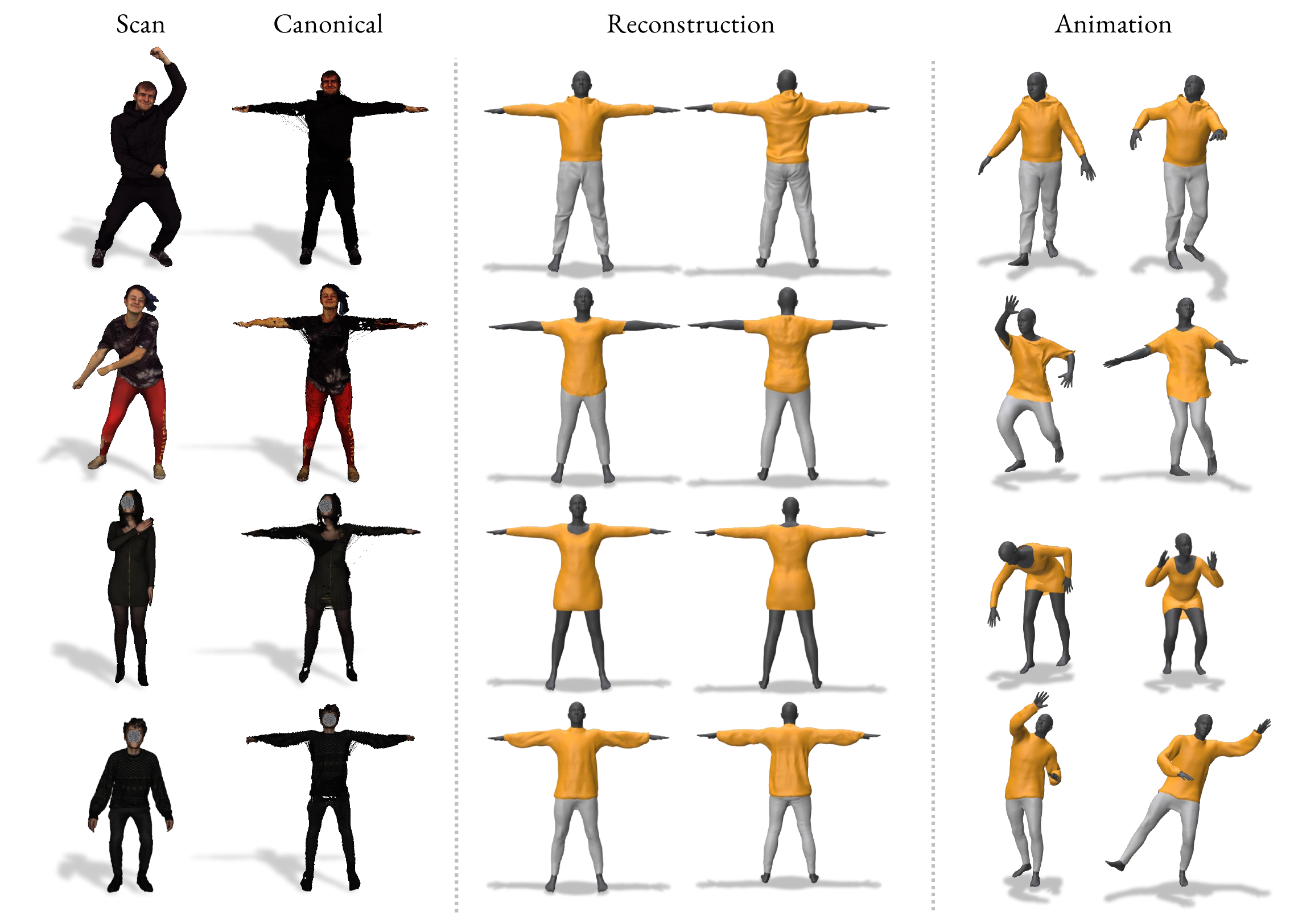}
    \captionof{figure}{\small{Garment reconstruction from 3D scans. We propose a pipeline to extract clothes from a posed scan, reconstruct them using Design2Cloth and animate them. Using Mask Encoder as a cloth prior we are able to reconstruct corrupted garments with holes and irregular structure.}
    \label{fig:Cloth_Repose}}
\end{figure*}
\begin{table}
\begin{center}
\caption{Cloth Reconstruction Error from single image on CustomHumans Dataset \cite{ho2023custom}.}
\label{tab:customhuman_reconstruction}
\vspace{-0.2cm}
\resizebox{\linewidth}{!}{
\begin{tabular}{l|c|c|c|c}
        \hline
Method   &  SMPLicit  &  ClothWild & DrapeNet &\textbf{Proposed} \\ \hline
CD ($\times 10^{-2}$) $\downarrow$ & 0.40 & 0.62 & 1.14 & \textbf{0.12}  \\ \hline \hline
\end{tabular}}
\end{center}
\vspace{-0.5cm}
\end{table}
\begin{figure}[!h]
    \centering
    \includegraphics[width=\linewidth]{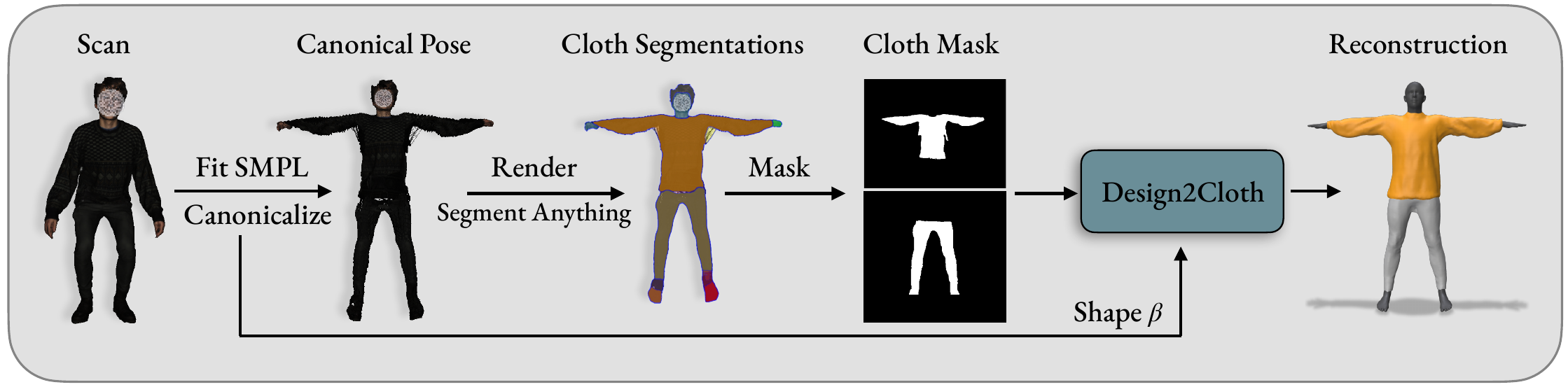}
    \captionof{figure}{\small{The proposed pipeline used to reconstruct clothes from corrupted 3D scans.}
    \label{fig:repose_pipeline}}
\end{figure}
\subsection{Reconstruction and Animation of Raw Scans}
In addition to its user-friendly nature, the proposed mask cloth representation facilitates the reconstruction of corrupted cloth data. As an application case we examined the case of reconstructing and animating raw scans with partial observations. In particular, using a simple LBS we canonicalized posed scans from the CustomHumans dataset \cite{ho2023custom} and rendered them to obtain their corresponding 2D visibility masks. The visibility masks are then mapped to the real garment distribution using the Mask Encoder $\mathbf{E}_m$.
It is noteworthy that, even though both the masks and the canonicalized garments may contain holes and irregular topologies, the cloth reconstructions generated using Design2Cloth are devoid of such artifacts. Finally, the reconstructed meshes can then be animated using any off-the-shelf draping method \cite{Grigorev_2023_CVPR,De_Luigi_2023_CVPR}. The overall 3D scan reconstruction and animation pipeline is depicted in \cref{fig:repose_pipeline}. 
\cref{fig:Cloth_Repose} illustrates the garment reconstructions and animations on CustomHumans (top two rows) and DigitalMe (bottom two rows) datasets. The proposed method remains unaffected by the artifacts created during the inverse blend skinning and can accurately reconstruct the 3D garments from the corrupted canonical scans. Under this setting, Design2Cloth can be used to restore and rectify garments from corrupted and occluded clothed human scans.

\section{Conclusion}
In this study we introduce Design2Cloth, a high fidelity 3D garment generative model that significantly develops current state-of-the-art generative models. Trained on a collected large scale real-world dataset, composed of more than 2,000 garments from 2,010 distinct identities, the proposed method is able to produce highly diverse and realistic garments. Significantly, in contrast to previous methods, Design2Cloth provides a user-friendly technique for garment design from simple 2D masks. In addition, the proposed pipeline is fully differentiable, providing a plug-and-play solution to several inverse problems, including cloth reconstruction from single images and scans. We believe that Design2Cloth can not only aid garment learning using synthetic clothes sampled from latent space, but also abet the realm of 3D design. \\

\noindent\textbf{Acknowledgements.} S. Zafeiriou was supported by EPSRC Project DEFORM (EP/S010203/1) and GNOMON (EP/X011364). R.A. Potamias was supported by EPSRC Project GNOMON (EP/X011364).
{
    \small
    \bibliographystyle{ieeenat_fullname}
    \bibliography{main}
}


\end{document}